\documentclass[journal]{IEEEtran}
\usepackage{graphicx}
\usepackage{multirow}
\usepackage{booktabs}
\usepackage{amsmath}
\usepackage{amssymb}
\usepackage[switch]{lineno}
\usepackage{amsfonts}
\usepackage{pifont}%
\usepackage{colortbl}
\usepackage{color,xcolor}
\usepackage{flushend}
\usepackage{xcolor, soul}
\definecolor{shadecolor}{rgb}{0.92,0.92,0.92}  
\usepackage{framed}  
\usepackage{cite}  
\ifCLASSINFOpdf
\else
\fi

\usepackage{xcolor}
\usepackage{soul}
\usepackage{hyperref}
\begin{document}
	
	\definecolor{Seashell}{RGB}{250, 250, 0} 
	\definecolor{Firebrick4}{RGB}{0, 0, 0}
	
	\newcommand{\code}[1]{
		\begingroup
		\sethlcolor{Seashell}
		\textcolor{Firebrick4}{\hl{#1}}
		\endgroup
	}
	
	\definecolor{Seashell_pol}{RGB}{0, 250, 0} 
	\definecolor{Firebrick4_pol}{RGB}{0, 0, 0}
	\def\ie{\textit{i.e.}}
	\def\eg{\textit{e.g.}}
	\def\etal{\textit{et al.}}

	\newcommand{\pol}[1]{
		\begingroup
		\sethlcolor{Seashell}
		\textcolor{Firebrick4}{\hl{#1}}
		\endgroup
	}
	
	%
	\title{GSmoothFace: Generalized Smooth Talking Face Generation via Fine Grained 3D Face Guidance}
	%
	%
	%
	\author{Haiming Zhang, Zhihao Yuan, Chaoda Zheng, Xu Yan, Baoyuan Wang, Guanbin Li, \\ Song Wu, Shuguang Cui,~\IEEEmembership{Fellow,~IEEE}, and Zhen Li$^{\star}$
		\thanks{Haiming Zhang, Zhihao Yuan, Chaoda Zheng, Xu Yan are with The Future Network of Intelligence Institute, School of Science and Engineering, and The Chinese University of Hong Kong (Shenzhen).
        E-mail: \{haimingzhang, zhihaoyuan, chaodazheng, xuyan1\}@link.cuhk.edu.cn,}
        \thanks{Shuguang Cui, Zhen Li are with School of Science and Engineering, The Future Network of Intelligence Institute, and The Chinese University of Hong Kong (Shenzhen)
        E-mail: \{shuguangcui, lizhen\}@cuhk.edu.cn,}
        \thanks{Baoyuan Wang is with Xiaobing.ai. Email: zjuwby@gmail.com,}
		\thanks{Guanbin Li is with Sun Yat-sen University, China. Email: liguanbin@mail.sysu.edu.cn,}
      \thanks{Song Wu is with Department of Urology, South China Hospital, Health Science Center, Shenzhen University, Shenzhen. Email: wusong@szu.edu.cn.}
	}
	\maketitle
	
	\begin{abstract}
		Although existing speech-driven talking face generation methods achieve significant progress, they are far from real-world application due to the avatar-specific training demand and unstable lip movements.
        To address the above issues, we propose the GSmoothFace, a novel two-stage generalized talking face generation model guided by a fine-grained 3d face model, which can synthesize smooth lip dynamics while preserving the speaker's identity.
        Our proposed GSmoothFace model mainly consists of the Audio to Expression Prediction (A2EP) module and the Target Adaptive Face Translation (TAFT) module.
        %
        %
        %
        Specifically, we first develop the A2EP module to predict expression parameters synchronized with the driven speech. 
        It uses a transformer to capture the long-term audio context and learns the parameters from the fine-grained 3D facial vertices, resulting in accurate and smooth lip-synchronization performance.
        Afterward, the well-designed TAFT module, empowered by Morphology Augmented Face Blending (MAFB), takes the predicted expression parameters and target video as inputs to modify the facial region of the target video without distorting the background content.
        The TAFT effectively exploits the identity appearance and background context in the target video, which makes it possible to generalize to different speakers without retraining. 
        Both quantitative and qualitative experiments confirm the superiority of our method in terms of realism, lip-synchronization, and visual quality.
        See the project page for code, data, and request pre-trained models: \href{https://zhanghm1995.github.io/GSmoothFace}{https://zhanghm1995.github.io/GSmoothFace}.
		
	\end{abstract}
	
	\begin{IEEEkeywords}
		Deep Learning, Talking Face Generation, Transformer, Generative Adversarial Model
	\end{IEEEkeywords}
	
	%
	\IEEEpeerreviewmaketitle
	
	\section{Introduction}
	
	
	\IEEEPARstart{S}{}peech-driven talking face generation aims to synthesize realistic portrait videos with lip movements synchronizing with arbitrary speech input. It has become an important technique since it has a wide range of applications in digital human animation~\cite{cudeiro2019capture, Richard_2021_ICCV_MeshTalk, zhou2018visemenet}, visual dubbing~\cite{xie2021towards}, virtual video conferencing~\cite{Wang_2021_CVPR_Conferencing}, and entertainment~\cite{MakeItTalk} fields. 

Previous works can be roughly divided into 2D-based and 3D-based methods.
The 2D-based methods usually formulate the audio-driven talking face generation problem as a conditional Generative Adversarial Networks (cGANs)~\cite{jamaludin2019you, kr2019towards, vougioukas2020realistic, prajwal2020lip, zhou2019talking, eskimez2021speech}.
However, due to the implicit supervision for audio-to-lip movement mappings learning and several inherent limitations of GANs, such as unstable training and mode collapse, they produce preliminary results with low image quality and unsatisfied lip-synchronization.
Other works~\cite{MakeItTalk, chen2019hierarchical, xie2021towards, Lahiri_2021_LipSync3D} utilize facial landmarks as an intermediate representation to bridge the distinct audio and visual modalities.
Nevertheless, most of these approaches struggle in generating vivid and natural talking face videos with delicate lip details, owing to the sparsity of the facial landmarks.
There are also some works exploring the few-shot and one-shot talking face generation via flow-based motion fields~\cite{Zhang_2021_CVPR_flow, ren2021pirenderer, wang2021one}.
Despite realizing a certain degree of generalizability, they suffer from the distorted background and unnatural face shapes caused by the limited context information from a few images.

\begin{figure}[t]
 \begin{center}
  \includegraphics[width=0.9\linewidth]{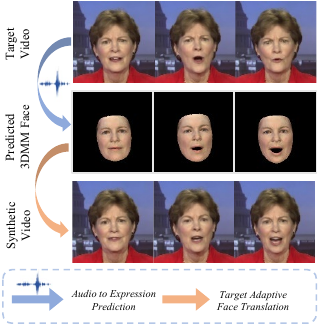}
 \end{center}
 \caption{Illustration of our generalized smooth talking face generation model via fine-grained 3DMM face guidance. Taking the target video and driven audio as inputs, our model predicts the lip-synced expression parameters via the Audio to Expression Prediction (A2EP) module. Next, we apply the Target Adaptive Face Translation (TAFT) module to synthesize the video with synchronized lips, while keeping the head pose and background the same as the target video.}
 \label{fig:fig1}
 \vspace{-.3cm}
\end{figure}

\section{Introduction}
\label{sec:intro}


Recently, the 3D-based methods~\cite{yi2020audio_personalized, thies2020NVP, zhang2021facial, ren2021pirenderer} relying on the 3D Morphable Model (3DMM)~\cite{egger20203dmm} obtain more attention, since it possesses the ability of 3D-aware modeling and parameters disentanglement.
%
However, almost all of these approaches ignore the fact that the ground truth of 3D face parameters estimated from existing state-of-the-art reconstruction algorithms~\cite{dengAccurate3DFace2020} are inconsistent and noisy~\cite{yin2022styleheat, wang2022lip}.
Thus they are hard to learn long-time audio-to-lip mappings for synthesizing stable and accurate lip movements, although applying the sliding window smoothing strategy~\cite{yin2022styleheat}.
Besides, these methods either require retraining for each new speaker~\cite{thies2020NVP, yi2020audio_personalized, zhang2021facial} or fail to recover the background region~\cite{gupta2023towards}.
More recently, the Neural Radiance Fields (NeRF) based talking head methods~\cite{guo2021ad_nerf, yao2022dfa, ye2023geneface} have achieved great progress since NeRF could render high-fidelity images with rich details such as hair and wrinkles. However, the issues of time-consuming training and poor generalizability hinder practical applications.

To address the above issues, we propose a two-stage talking face generation framework, namely \textbf{GSmoothFace}. It contains an Audio to Expression Prediction (A2EP) module and a Target Adaptive Face Translation (TAFT) module. These modules can generate smooth lip movements and high-fidelity talking face videos with identity preservation. They can also generalize to unseen speakers without retraining.
Specifically, our A2EP consists of three components: an audio encoder, an expression encoder, and a weighted multi-head decoder. The audio encoder is based on transformer architecture and can capture long-term audio features, while most previous works consider relatively short-term audio context dependencies.
The decoder predicts face expressions in fine-grained facial vertices with more attention to the mouth region. Then the TAFT module exploits the estimated expression parameters to enerate accurate and smooth lip movements.
The TAFT module takes three inputs: the predicted expression parameters, the target video, and the original 3DMM parameters reconstructed from it. It uses a Morphology Augmented Face Blending (MAFB) module to obtain blended images that retain accurate mouth shapes but unreal textures.
Finally, the blended images and reference images from the target video are fed to a generator to synthesize talking face videos with promising identity-preserving performance without introducing distortions and artifacts (see Fig.~\ref{fig:fig1}).

Extensive experimental results on widely-used VoxCeleb and HDTF datasets demonstrate the superiority of our proposed method and the effectiveness of several key components. 
Overall, our main contributions are three-fold:
\begin{itemize}
\setlength{\itemsep}{0pt}
\setlength{\parsep}{0pt}
\setlength{\parskip}{0pt}
\item We propose GSmoothFace, a simple yet effective two-stage generalized talking face generation framework. It leverages the fine-grained 3D face model to generate realistic talking face videos with promising lip synchronization and identity-preserving performance.
\item We introduce a transformer-based Audio to Expression Prediction (A2EP) module, which considers the long-term audio context dependencies and supervises the training process in an explicit way.
\item We design a Target Adaptive Face Translation (TAFT) module, which can blend the fine-grained rendered face image that already has correct lip motions with the target video to synthesize a realistic talking face video with very few artifacts. And it can generalize to unseen identity without retraining.
\end{itemize}

\begin{figure*}[t]
\begin{centering}
    \center\includegraphics[width=\linewidth]{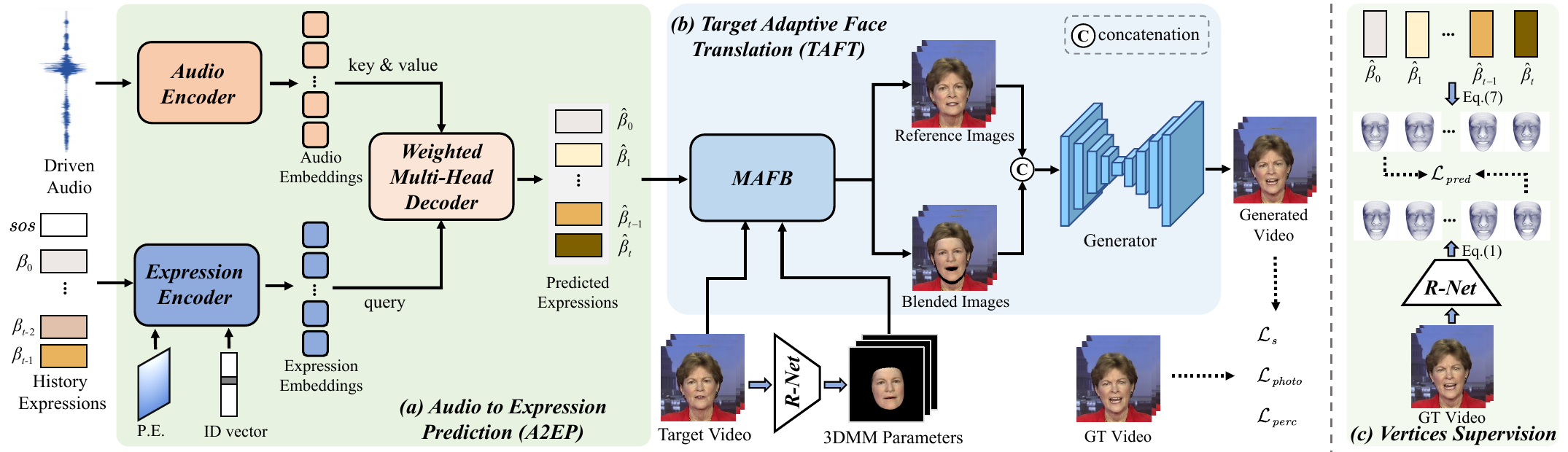}
    \caption{Illustration of our proposed \textbf{GSmoothFace} framework, which mainly consists of the Audio to Expression Prediction (A2EP) and Target Adaptive Face Translation (TAFT) modules. \textbf{(a) A2EP.} Given driven audio and history expressions with identity vectors as well, the predicted expressions are obtained via a weighted attention transformer in an auto-regressive manner (Sec.~\ref{sec:audio-to-expression}).
     Note that the expressions are supervised in the fine-grained vertices aspects, as \textbf{(c)} shows.
     \textbf{(b) TAFT}. Taking the predicted expression parameters and original 3DMM parameters reconstructed by a pre-trained state-of-the-art method R-Net (Sec.~\ref{sec:face_modeling}) from the target video as inputs, the blended images are obtained via the simple but effective fully differential Morphology Augmented Face Blending (MAFB) module (Sec.~\ref{sec:target_adaptive_face_translation}). Afterward, the blended images, combined with reference images containing the identity information, are used to synthesize a smooth photo-realistic talking face video by a generator network. Which can be generalized to unseen speakers without retraining.
    }
    \label{fig:framework}
\end{centering}	
\end{figure*}
	
\section{Related Work}
\label{sec:related}
\subsection{2D-based Talking Face Generation.}
A large number of 2D-based methods~\cite{kr2019towards, prajwal2020lip, park2022synctalkface, wen2020photorealistic, liang2022expressive} have been proposed to synthesize talking face videos with conditional generative adversarial networks.
Several works~\cite{chen2019hierarchical, MakeItTalk, lu2021live, ji2021EVP} leverage the 2D facial landmarks as the intermediate representation to alleviate the challenges of audio-to-image translation.
There are also some works focusing on the few-shot~\cite{zakharov2019few} and one-shot~\cite{wang2021audio2head, wang2021one, Wang_2021_CVPR_Conferencing, Zhang_2021_CVPR_flow, ren2021pirenderer, yin2022styleheat, ji2022eamm} talking head generation by means of meta-learning~\cite{zakharov2019few} or motion field flow~\cite{wang2021one, Zhang_2021_CVPR_flow, yin2022styleheat}. 
%
Although achieving generalization capacity, it is inevitable to bring many obvious artifacts in the synthesized videos, \textit{e.g.}, the distorted background and unnatural face shapes, due to the background occlusions.
In comparison, several video-based talking face generation methods~\cite{ji2021EVP, prajwal2020lip, park2022synctalkface} tend to obtain results with few distorted regions, since they keep most
parts of the target videos unchanged while focusing on editing mouth region only.
However, since these 2D-based methods do not consider the 3D structure inherence of human heads, the smoothness and naturalness of the synthesized talking videos are inferior to the 3D-based methods.

\subsection{3D-based Talking Face Generation.}
More and more works exploit the attractive 3DMM~\cite{thies2020NVP, yi2020audio_personalized, zhang2021facial, song2022everybody, Lahiri_2021_LipSync3D, lv2022generating, huang2022audio} and FLAME~\cite{SiggraphAsia2017_FLAME, cudeiro2019capture, richard2021meshtalk, faceformer2022} models to improve the 2D-based talking face generation, owing to their effective 3D structure modeling and superior parameters disentanglement abilities. 
Despite all of them utilizing the 3DMM as the intermediate representation, they are different in the ways of how to learn the audio to 3DMM mapping and how to obtain the talking face videos from 3DMM.
~\cite{zhang2021facial}, \cite{thies2020NVP} and \cite{lv2022generating} leverage the simple temporal-aware convolution operations to predict the 3DMM parameters directly from the audio waveform, and \cite{yi2020audio_personalized, song2022everybody} further improve the audio encoder by using the inefficient RNN and LSTM. Although leveraging the transformer as the encoder, \cite{huang2022audio} still uses the traditional acoustic feature to represent the audio waveform.  As regards the rendering part, they either leverage the GANs to translate the rendered face images into realistic ones containing background information~\cite{thies2020NVP, yi2020audio_personalized, lv2022generating, song2022everybody} while losing the generalizability, or implicitly embed the predicted 3DMM parameters into latent codes for image editing~\cite{zhang2021facial, huang2022audio} but with more artifacts.
%
%

\subsection{NeRF-based Talking Face Generation.}
More recently, NeRF-based methods have received broad attention. 
AD-NeRF~\cite{guo2021ad_nerf} is the first to apply NeRF in the area of talking head generation. DFA-NeRF~\cite{yao2022dfa} tries to capture synchronized lip motion and personalized attributes simultaneously via a disentangled face attributes neural radiance field. To reduce the required training data scale, DFRF~\cite{shen2022derf} learns a dynamic facial radiance field conditioned on the reference images. 
GeneFace~\cite{ye2023geneface} exploits a generative audio-to-motion model and regards the NeRF as a renderer to obtain high-quality images.
Although showing great potential in talking face generation, the lip-synchronization performance of current NeRF-based methods is still far away from non-NeRF methods. 
And the demand of time-consuming retraining for each speaker also hinders the practicability.

\section{Method}
\label{sec:data}
Given a target video $\mathcal{V}_t$ of a speaker and an arbitrary speech audio $\mathcal{X}$, the speech-driven talking face generation aims to synthesize a video $\mathcal{\hat{V}}$ saying this speech with correct lip synchronization while preserving the speaker's identity. 
As discussed above, in our setting, we retain the head poses and background the same as the target video.
As shown in Fig.~\ref{fig:framework}, our method consists of two key components, \ie, the Audio to Expression Prediction (A2EP) and Target Adaptive Face Translation (TAFT). 
The A2EP module takes the driven audio $\mathcal{X}$ and historical expressions as inputs to predict the expression parameters $\hat{\beta}$ in an auto-regressive manner (Sec.~\ref{sec:audio-to-expression}). 
Note that we supervise the predictions in the fine-grained facial vertices with more attention on the mouth region, which alleviates the inevitable 3D face reconstruction errors between consecutive frames, and achieve more accurate and smooth lip movements. 
Then the TAFT module is responsible to modify the face regions in the target video with synced lip movements with driven audio.
Specifically, the Morphology Augmented Face Blending (MAFB) module utilizes the target video and its reconstructed 3DMM parameters, together with the predicted expression parameters $\hat{\beta}$ to obtain blended images with correct lip shapes (Sec.~\ref{sec:target_adaptive_face_translation}).
After that, a simple UNet-like generator is leveraged to generate the photo-realistic talking face video $\mathcal{\hat{V}}$ (Sec.~\ref{sec:target_adaptive_face_translation}), by talking the blended images and reference images $\mathcal{V}_{ref}$ as inputs.

\subsection{3D Face Modeling} \label{sec:face_modeling}
We first have a brief review of the 3DMM face modeling and then introduce the audio to expression prediction algorithm based on it. 
In 3DMM, the 3D shape $\mathbf{S}$ and texture $\mathbf{T}$ of a face  and texture are parameterized by an affine model~\cite{egger20203dmm, blanz1999morphable}:
\begin{equation}
\begin{aligned}
    \mathbf{S} &= \mathbf{S}(\alpha, \beta) = \mathbf{\bar{S}} + \mathbf{B}_{id}\alpha +\mathbf{B}_{exp}\beta, \\
    \mathbf{T} &= \mathbf{T}(\delta) = \mathbf{\bar{T}} + \mathbf{B}_{tex}\delta,
\end{aligned}
\end{equation}
where $\mathbf{\bar{S}}$ and $\mathbf{\bar{T}} \in \mathbb{R}^{3N}$ denote the average face shape and texture with $N$ vertices.
$\mathbf{B}_{id}$, $\mathbf{B}_{exp}$ and $\mathbf{B}_{tex}$ are the PCA bases of identity, expression, and texture respectively. And $\alpha$, $\beta$ and $\delta$ are the corresponding coefficients vectors for generating a 3D face.
To preserve the head pose variance, the 3D face poses parameters $p$ that are represented by rotation $r \in SO(3)$ and translation $t \in \mathbb{R}^3$ are also included.

Following previous work~\cite{zhang2021facial, ren2021pirenderer, yi2020audio_personalized}, we leverage a pre-trained state-of-the-art 3D face reconstruction model~\cite{dengAccurate3DFace2020} (the \textbf{R-Net} in Fig.~\ref{fig:framework}), to regress all of this 3DMM parameters $\mathbf{x}=(\alpha, \beta, \delta, p)$ from a face image.
%
And in this case, we can get a fine-grained 3D face mesh containing $N=35709$ vertices, which is dense enough to capture various facial expressions, especially the subtle lip motions. 
We apply the illumination and perspective camera models from~\cite{dengAccurate3DFace2020} as well to render the 2D analytic image from the 3D face model.
Besides, we also generate a binary mask image indicating the face region through the 3D face mesh projection operation~\cite{dengAccurate3DFace2020}, which is used in our MAFB module (Sec.~\ref{sec:target_adaptive_face_translation}).

\begin{figure}[t]
 \begin{center}
  \includegraphics[width=\linewidth]{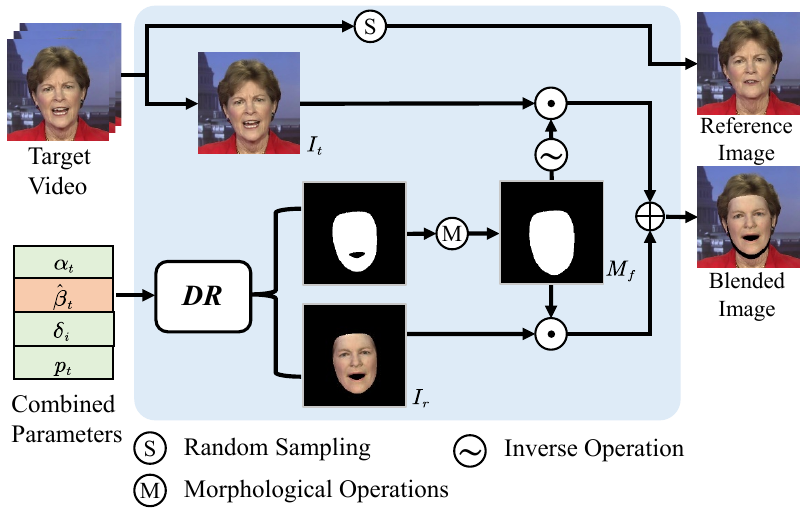}
 \end{center}
 \caption{Illustration of Morphology Augmented Face Blending module (MAFB). Here, DR denotes the differentiable renderer, $\odot$ and $\oplus$ represent the element-wise multiplication and addition operation, respectively. Other symbols are explained at the bottom. }
 \label{fig:face_blending_module}
\end{figure}

\subsection{Audio to Expression Prediction (A2EP)}\label{sec:audio-to-expression}
\noindent \textbf{Temporal-aware Audio Encoder.} Given a raw audio waveform $\mathcal{X}$, our temporal-aware audio encoder produces a sequence of discrete audio embeddings $\mathbf{F}_a=\{a_t\}_{t=0}^{T-1}$, where $a_t \in \mathbb{R}^\mathbf{d}$ and $T$ denotes the number of frames in the corresponding video. 
Here, we employ a pre-trained wav2vec 2.0~\cite{baevski2020Wav2Vec} as our audio encoder, which is one of the popular masked language modeling approaches~\cite{devlin2018bert} and pre-trained in large-scale speaking datasets in a self-supervised scheme. 
Therefore, it's beneficial for generating smooth lip dynamics since it not only considers local audio contextual information but also incorporates the information from other time steps. 
The original wav2vec 2.0 model outputs audio sequence embeddings with a frequency that is different from the video frame rate, so we add an additional linear interpolation layer for resampling the audio features along the temporal dimension, to ensure the alignment of audio embeddings and the following expression embeddings. 
During training, we choose to finetune the encoder to well adapt our task. More details can be seen in our supplementary materials.

\noindent \textbf{Expression Encoder.} The expression encoder accepts the historical outputs of the decoder to extract expression embeddings in the next time stamp.
Concretely, we first embed the history of predicted expression parameters sequence into high dimensional expression features through a trainable linear projection layer, obtaining $\mathbf{F}_{e}^{\prime}=\{e_t^{\prime}\}_{t=0}^{T-1}$, where $e_t^{\prime} \in \mathbb{R}^\mathbf{d}$ also has $\mathbf{d}$ dimension. Note that we set a zero-vector as the sign of start (sos in Fig.~\ref{fig:framework}) since we do not know the expressions at the first frame.

Afterward, inspired by VOCA~\cite{cudeiro2019capture}, we also embed the one-hot vector of speaker identity to a high-dimensional vector $\mathrm{v}_n$ via a linear projection layer and add it to the expression feature tokens, which are used to distinguish the different speaking styles to facilitate the training process.
Besides, the commonly-used positional encoding (P.E.) strategy~\cite{vaswani2017attention} is utilized to introduce temporal order information.
Overall, the expression embedding process can be represented as:
\begin{equation}
\begin{aligned}
    e_t^{\prime} &= \mathbf{FC}(\hat{\beta}_{t-1}) + \mathrm{v}_n \\
    \tilde{e}_t &= e_t^{\prime} + \mathbf{PE}(t) \\
    \mathbf{\tilde{F}}_e &=\{\tilde{e}_t\}, 0\leq t < T-1,
    \label{eq_cross_attension}
\end{aligned}
\end{equation}
where $\mathbf{FC}(.)$ and $\mathbf{PE}(.)$ denote the linear projection layer and positional embedding respectively.
Finally, the expression embeddings sequence $\mathbf{\tilde{F}}_e$ is fed into the multi-head self-attention layers to capture the temporal contextual information, resulting in temporal-aware expression embeddings $\mathbf{F}_e$.

\noindent \textbf{Transformer Decoder.}
As shown in Fig.~\ref{fig:framework}, the transformer decoder takes the audio embeddings and the expression embeddings as inputs, to predict the expression parameters synced with the driven audio.

As the core function of the transformer decoder, cross-attention is designed for inter-modality interaction to obtain embedded cross-modal features. 
Different from the self-attention, now the expression embeddings $\mathbf{F}_e$ are linearly transformed into query matrix $Q_e$, and the $\mathbf{F}_a$ is transformed into the key matrix $K_a$ and value matrix $V_a$. 
In our experiments, we found that it is not very easy for the vanilla transformer to regress subtle variations among adjacent frames since the video sequences have high frequency (\eg 25 FPS in our settings). 
To alleviate this issue, we introduce a linear bias in cross-attention layers to constrain the dependencies range of cross-attention~\cite{press2021_alibi,faceformer2022}:
\begin{equation}
    \mathrm{BiasAtt}(Q_e,K_a,V_a) = \mathrm{softmax}(\frac{Q_eK_a^T}{\sqrt{\mathbf{d}}} + B_{a,e})V_a,
    \label{eq_cross_attension}
\end{equation}
where $B_{a,e}$ is a static, non-learned bias matrix and can be represented as:
\begin{equation}
\small
\begin{aligned}
    B_{a,e}(i, j)&= \begin{cases}0, & \mathrm{max}(i-\sigma_1,0) \leq j<\mathrm{min}(i+\sigma_2, T-1) \\
     -\infty, & \text { otherwise }\end{cases},
\end{aligned}
\label{eq:bias_matrix}
\end{equation}
where $0 \leq i < t, 0 \leq j < T-1$ are the matrix index value, $\sigma_1, \sigma_2$ control the forward and backward bias steps respectively.
Similarly, we also utilize the multi-head attention mechanism in the cross-attention. Finally, a feed-forward layer is used to transform the hidden state to the final expression parameters sequence $\hat{\beta}_t$.

\noindent \textbf{Loss Function.} 
In our experiments, we found that directly supervising the expression parameters as previous works~\cite{zhang2021facial,yi_zhangjuyong_2020,thies2020NVP} in our design cannot obtain a satisfactory result due to inevitable reconstruction errors~\cite{dengAccurate3DFace2020} between consecutive frames of each video. 
%
%
To this end, we supervise the training in the dense facial vertices aspect and make two simple but effective modifications to alleviate these issues.

In practice, we first compute a mean identity coefficient $\hat{\alpha}$ for each speaker and use it to obtain an approximate template face vertices:
\begin{equation}
    \mathbf{S}_{temp} = \mathbf{\bar{S}} + \mathbf{B}_{id}\bar{\alpha}.
\end{equation}
Then the pseudo ground truth $\mathbf{S}_{gt}^t \in \mathbb{R}^{3N}$ and our predicted facial vertices $\mathbf{S}_{pred} \in \mathbb{R}^{3N}$ in frame $t$ can be defined as
\begin{equation}
\begin{aligned}
    \mathbf{S}_{gt}^t = \mathbf{S}_{temp} + \mathbf{B}_{exp}\beta_t, \\
\end{aligned}
\end{equation}
\begin{equation}
\begin{aligned} 
    \mathbf{S}_{pred}^t = \mathbf{S}_{temp} + \mathbf{B}_{exp}\hat{\beta}_t. \\
\end{aligned}
\end{equation} 
Besides, we generate a binary vector $M \in \mathbb{R}^N$ by using facial vertices of 3DMM mean face under threshold conditions. 
In $M$,  we use $1$ to represent the facial vertices indices belonging to the lower mouth region, others all $0$. 
Thus, we can make the network gain more attention to the prediction of audio-related vertices.
Finally, we apply the MSE (Mean Squared Error) loss to minimize the error between the predicted facial vertices and the pseudo ground truth as
\begin{equation}
\begin{aligned}
    \mathcal{L}_{pred} = &\frac{1}{T}\sum_{t=0}^{T-1}(\lambda_m \|M \odot (\mathbf{S}_{pred}^t-\mathbf{S}_{gt}^t)\|_2 + \\
     &\|(1-M) \odot (\mathbf{S}_{pred}^t-\mathbf{S}_{gt}^t) \|_2),
\end{aligned}
\end{equation}
where $\odot$ denotes element-wise multiplication, $T$ is the length of the video. And we set $\lambda_m>1$ to give a larger weight to the mouth area.

\subsection{Target Adaptive Face Translation (TAFT)}\label{sec:target_adaptive_face_translation}
We propose the TAFT module instead of most of the existing works that depend on long videos of specific speakers to completely synthesize the full images to ``remember" the background information, resulting in poor generalizability.
The proposed TAFT not only retains the readied speech-driven 3D facial expressions predicted by A2EP but also avoids the background distortions and artifacts caused by full image re-generation.

\noindent \textbf{Morphology Augmented Face Blending (MAFB).}
%
As Fig.~\ref{fig:face_blending_module} illustrated, the MAFB module first obtains the combined 3DMM parameters by replacing the original expression parameters reconstructed from target video frames with the predicted ones by A2EP. 
Then, the differentiable renderer (DR) takes these combined 3DMM parameters as input to obtain the 3DMM rendered face image $I_r$. 
Meanwhile, we generate the corresponding binary face mask image from the depth information of the 3D face mesh as a byproduct without much further effort. 
Note that there is a black hole around the mouth region since there are no vertices in the inner mouth in 3DMM.
In order to blend the 3DMM rendered face and the target image, we design a morphology-based mask augmentation method.
Specifically, we first apply the morphological closing operation to erase the black unmasked inner mouth region. After that, we randomly employ morphological dilation or erosion operations with different pre-defined kernel sizes, resulting in face masks $\mathit{M}_f$ with various scales, which is helpful to adapt to the changes of the face shapes caused by different expressions.
Finally, we obtain the blended image $I_b$ by using pixel-wise operations, formulated as:
\begin{equation}
\begin{aligned}
    I_b = I_r \odot \mathit{M}_f + I_t \odot (1-\mathit{M}_f),
\end{aligned}
\label{equ:Lrender}
\end{equation}
where $I_t$ is the frame-by-frame image from the target video.

After that, a generator synthesizes the final image $\hat{I}_t$ by concatenating $I_b$ and a randomly sampled image $I_\mathrm{ref}$ from the target video along channel dimension as inputs. 
Our generator is designed as an encoder-decoder architecture. The skip-connection layers are used in the generator to preserve the texture information from the reference image. More details about the generator can be found in the supplementary materials.

\begin{table*}[t]
  \centering
  \footnotesize
  \caption{The quantitative comparisons on VoxCeleb and HDTF datasets. ($\uparrow$): Higher is better. ($\downarrow$): Lower is better.}
  \label{tab:quan}
  \setlength\tabcolsep{3.5pt}
  \begin{tabular}{l|ccccc|ccccc}
    \toprule
     \multirow{2}[0]{*}{\textbf{Methods}} & \multicolumn{5}{c|}{VoxCeleb} & \multicolumn{5}{c}{HDTF} \\
     & \textbf{LMD} $\downarrow$ & \textbf{SSIM} $\uparrow$ & \textbf{PSNR} $\uparrow$ & \textbf{CPBD} $\uparrow$ & \textbf{AV conf.} $\uparrow$ & \textbf{LMD} $\downarrow$ & \textbf{SSIM} $\uparrow$ & \textbf{PSNR} $\uparrow$ & \textbf{CPBD} $\uparrow$ & \textbf{AV conf.} $\uparrow$ \\
    \midrule
    ATVG~\cite{chen2019hierarchical} & 1.892 & 0.578 & 21.092 & 0.082 & 5.120 & 1.670 & 0.679 & 22.205 & 0.071 & 6.381 \\
    Wav2Lip~\cite{prajwal2020lip} & 2.180 & 0.722 & 30.124 & 0.173 & 7.432 & 1.790 & 0.931 & 32.165 & 0.174 & \textbf{8.617} \\
    MakeItTalk~\cite{MakeItTalk} & 2.152 & 0.678 & 24.128 & 0.161 & 4.842 & 1.750 & 0.774 & 24.221 & 0.261 & 6.259 \\
    PC-AVS~\cite{zhou2021pose} & 1.480 & 0.582 & 20.102 & 0.182 & 7.134 & 2.440 & 0.466 & 13.405 & 0.144 & 8.143 \\
    \hline
    Ground Truth & 0.000 & 1.000 & inf & 0.252 & 7.893 & 0.000 & 1.000 & inf & 0.289 & 7.646 \\
    \textbf{Ours} & \textbf{1.280} & \textbf{0.858} & \textbf{38.852} & \textbf{0.212} & \textbf{7.345} & \textbf{0.820} & \textbf{0.987} & \textbf{43.292} & \textbf{0.268} & 7.314 \\
    \bottomrule
  \end{tabular}
\end{table*}

\noindent \textbf{Loss Function.} 
We supervise the training process of the TAFT module with a loss function including three parts: photo-metric loss, perceptual loss, and style loss.
The photo-metric loss measures the consistency between the real image and the synthesized
one in pixel level.
\begin{equation}
\begin{aligned}
    \mathcal{L}_{photo} = \|(\hat{I}_t - I_t) \|_1,
\end{aligned}
\end{equation}
The photo-metric loss could retain the average error in pixel level, but lack details. 
Therefore, similar to~\cite{siarohin2019fomm}, we calculate a perceptual loss on multi-scale resolutions by applying pyramid down-sampling on $I_t$ and $\hat{I}$, formulated as:
\begin{equation}
\begin{aligned}
    \mathcal{L}_{perc} = \sum_{i} \|(\phi_i(\hat{I}_t) - \phi_i(I_t)) \|_1,
\end{aligned}
\end{equation}
where $\phi_i$ denotes the activation map of the $i$-th layer of the pre-trained VGG-19 network~\cite{simonyan2014vgg}. The style loss $\mathcal{L}_s$ computes the statistic error between the VGG-19 activation map:
\begin{equation}
\begin{aligned}
    \mathcal{L}_s = \sum_{i} \|(G_{i}^{\phi}(\hat{I_t}) - G_{i}^{\phi}(I_t)) \|_1,
\end{aligned}
\end{equation}
where $G_{i}^{\phi}$ denotes the Gram matrix constructed from activation maps $\phi_i$.
Summing these three terms, we obtain the total loss function:
\begin{equation}
\begin{aligned}
    \mathcal{L} = \lambda_{photo} \mathcal{L}_{photo} + \lambda_p \mathcal{L}_{perc} + \lambda_s \mathcal{L}_s,
\end{aligned}
\end{equation}
where $\lambda_{photo}$, $\lambda_{perc}$, and $\lambda_s$ are weights for the photo-metric loss, perceptual loss, and style loss respectively.

 \begin{figure}[t]
 \begin{center}
  \includegraphics[width=\linewidth]{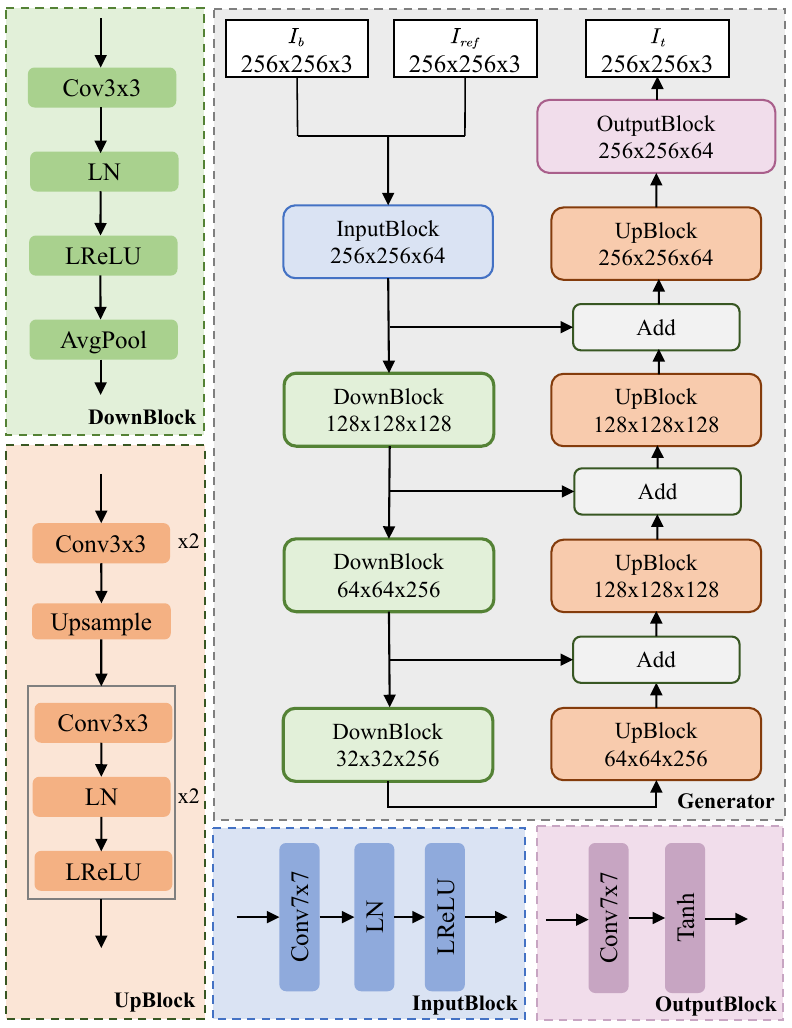}
 \end{center}
 \caption{Architectural details of our Generator in TAFT. Note that here we only demonstrate the network synthesizing 256$\times$256 face images, our generator can also generate images with 512$\times$512 resolution but with minor differences in the blocks' parameters. }
 \label{fig:generator}
\end{figure}

\begin{figure}[t]
 \begin{center}
  \includegraphics[width=\linewidth]{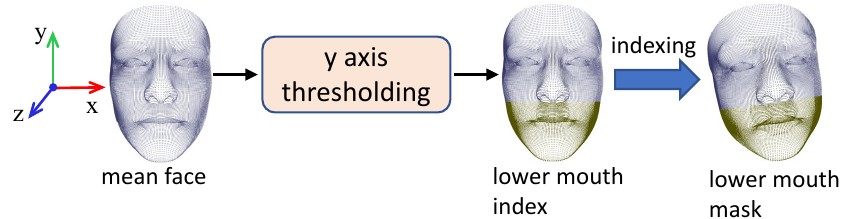}
 \end{center}
 \caption{The mouth mask generation process. Without further effort, we first obtain the vertex indices belonging to the lower mouth from the mean face of 3DMM. And then we exploit these indices on the 3D reconstructed faces of each speaker, to easily obtain the lower mouth mask of them.}
 \label{fig:mouth_mask_gen}
\end{figure}

\section{Experiments}
In this section, we conduct extensive experiments to validate the effectiveness of our proposed method, and the implementation details are also provided.

\subsection{Experimental Settings}
\noindent \textbf{Datasets.} We train and test our method on two popular in-the-wild audio-visual datasets, VoxCeleb~\cite{Nagrani17_voxceleb} and HDTF~\cite{Zhang_2021_CVPR_flow} dataset. Both datasets contain various speakers and different speaking styles, and offer Youtube links for downloading raw videos. 
VoxCeleb contains over 100K videos from 1251 speakers.
HDTF dataset consists of 362 videos with over 300 speakers, and the resolution of its original videos is 720\textit{P} or 1080\textit{P}, which is higher than the VoxCeleb dataset. 
Following~\cite{siarohin2019fomm}, we preprocess the VoxCeleb dataset by cropping faces with a fixed position window in each crop and filtering out sequences that have a resolution lower than $256\times256$. The remaining videos are resized to $256\times256$ preserving the aspect ratio. 
Due to some broken links, we finally obtained 7221 training videos and 244 test videos, with lengths varying from 64 to 1024 frames.
We also preprocess the HDTF dataset following~\cite{Zhang_2021_CVPR_flow}, obtaining videos with $512\times512$ resolution. 
Note that all videos are resampled in 25 FPS and the audio signals are resampled in 16 kHz.

\noindent \textbf{Evaluation Metrics.} \label{sec:Quantitative}
We conduct quantitative evaluations on several metrics widely used in previous approaches. 
To evaluate the image quality, we use The Peak Signal to Noise Ratio (PSNR), the Structure Similarity Index Measure (SSIM)~\cite{wang2004image}, and the perceptual-based no-reference objective image sharpness metric (CPBD)~\cite{narvekar2011no}.
For the mouth shape and lip synchronization evaluation, we utilize the Landmark Distance (LMD) metric~\cite{chen2018lip} and the Audio-Visual (AV) Offset and confidence score proposed in SyncNet~\cite{chung2016out}. 

\begin{figure*}[t]
\vspace{-.3cm}
 \begin{center}
  \includegraphics[width=\linewidth]{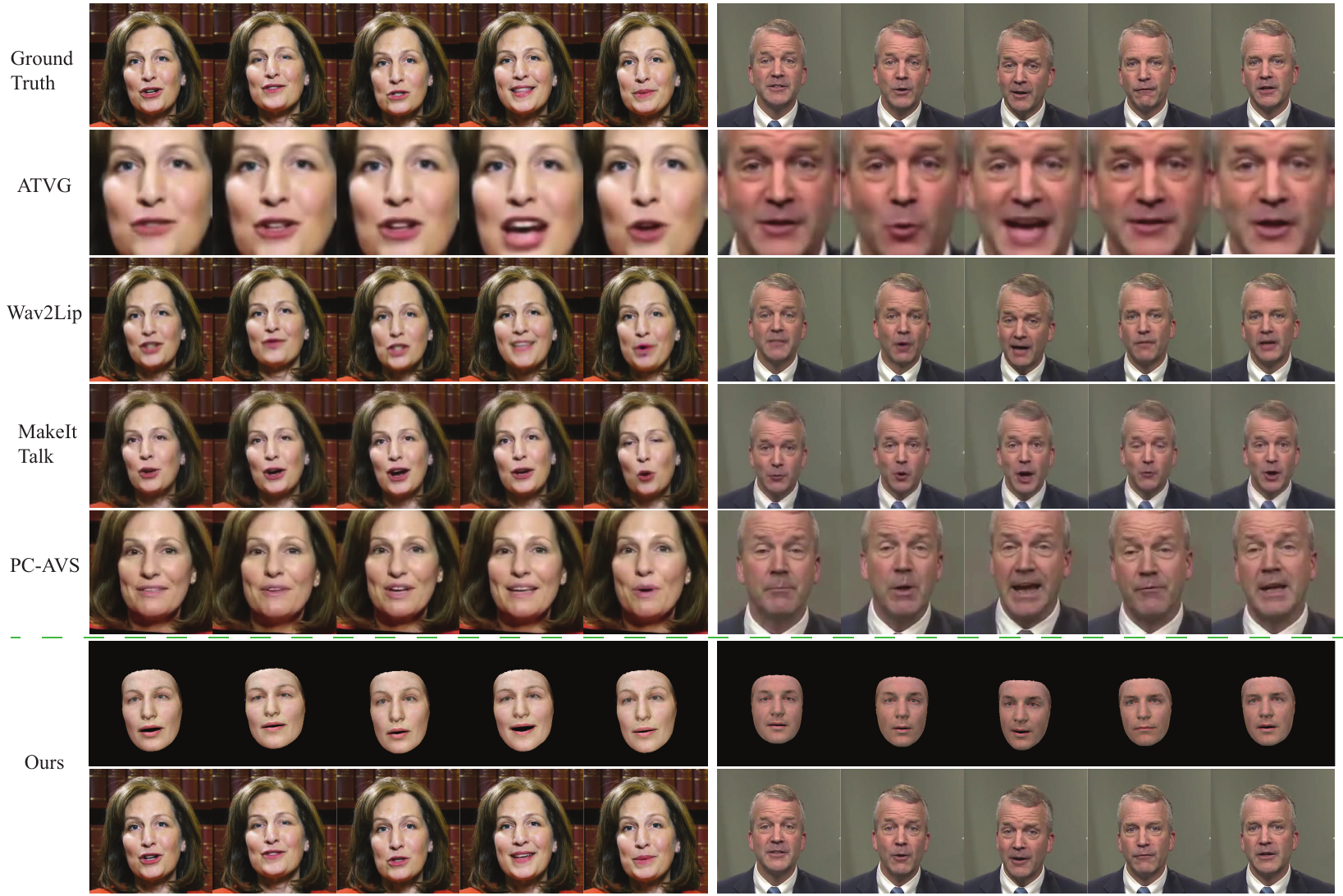}
 \end{center}
 \caption{Qualitative results on the HDTF dataset. (a) Qualitative comparisons between our method and state-of-the-art methods. Note that we also show our intermediate 3DMM rendered face images to demonstrate the ability of our method to synthesize accurate lip shapes. (b) Our generalized method can make different identities say the same speech while maintaining their own head poses and backgrounds with very few distortions and artifacts. 
 More results can be found in our supplementary materials.}
 \label{fig:quali}
 \vspace{-.3cm}
\end{figure*}
\subsection{Implementation Details}

\noindent \textbf{Architecture Implementation.} For the Temporal-aware Audio Encoder, we first resample the audio input to 16kHz and only retain a single channel data. 
Then, the multi-layer convolution operations are applied to get latent audio representations. 
After that, we use the wav2vec 2.0 transformer consisting of 12 identical layers with 768 dimensions and 12 attention heads, to extract temporal-aware contextual audio embeddings.
Note that before feeding the latent into the transformer encoder, we add a linear interpolation layer to interpolate the sequence length of audio features consistent with the corresponding video length.
Finally, a feed-forward layer transforms the 768-dimension features into $\mathbf{d}=1024$ dimension speech representations $\mathbf{F}_a$. 
During training, we initialize the parameters of the wav2vec 2.0 encoder with the weights pre-trained on 960 hours of unlabeled audio from LibriSpeech dataset. Afterward, we finetune all parameters for our task.

For the Expression Encoder, a linear layer projects the 64-dimension expression parameters into $\mathbf{d}=1024$ high dimensional features firstly.
Then we add the speaker identity embeddings, which are obtained from one-hot vectors by using another fully-connected layer. 
After that, we employ the multi-head self-attention with 4 heads and get the expression features embeddings $\mathbf{F}_e$ with $\mathbf{d}=1024$ dimension as well.

As the main component of the A2EP module, the cross-modal transformer decoder aims to accomplish the inter-modality interaction to obtain embedded cross-modal features for the prediction of expression parameters. 
The cross-modal transformer has one layer, and we employ a biased cross-attention mechanism to align audio features and expression features. 
In our experiments, we set $\sigma_1=0$ and $\sigma_2=1$ for the bias matrix in Eq.~\ref{eq:bias_matrix} in the main paper. 
And we also adapt 4 heads in the multi-head cross-attention, the dimension of the feed-forward is 2048. 
After obtaining the cross-modal embeddings, we utilize a linear projection layer to convert the embeddings to 64-dimension expression parameters as the network outputs. 
We initialize the last linear projection layer with 0s.

Fig.~\ref{fig:generator} illustrates the detailed structures of the generator in our Target Adaptive Face Translation module. 
Here, we use the $256\times256$ size image to show the different feature map scales, the network architecture is almost the same for the image with $512\times512$ size.

\noindent \textbf{Mouth Mask Generation}
When supervising the fine-grained 3D facial vertices generation, we emphasize more on vertices belonging to the mouth region, which is beneficial to subtle mouth motions. The original Basel Face Model (BFM)~\cite{paysan20093d} offers four segments of the face independently, including the nose, eyes, mouth, and others.
However, the mouth segment in BFM is too small to capture the mouth movements. 
This is because when speaking, not only does the mouth move but also the jaw and cheekbone are moving together.
Therefore, in this paper, we consider the region lowering than the nose for better supervision. 
However, it is not easy to obtain this region directly from each reconstructed 3D face model, as each face has its own facial shape and various head poses. 
Fortunately, we observe that the reconstructed 3D facial vertices have a fixed index order, and we can easily get the coordinate values of the mean face. 
Thus we extract the vertex indices belonging to the lower mouth region from the mean face, and then apply these indices to obtain the region for every reconstructed 3D face. 

Specifically, as shown in Fig.~\ref{fig:mouth_mask_gen}, we utilize the standard coordinate frame in the BFM mean face, where the x-axis and y-axis are right-ward and upward, respectively, and the z-axis is outward to construct a right-hand coordinate frame. 
Note that the origin point is located in the center of the nose. In order to get the lower mouth indices, we apply a threshold condition as follows:
\begin{equation}
    \mathcal{I}=\{i|V_{i}^{y}<y_m, \; \forall V_i \in \mathbf{\bar{S}}\},
\end{equation}
where $\mathbf{\bar{S}} \in \mathbb{R}^{N\times3}$ is the reshaped mean face vertices coordinates, $V_i$ and $V_{i}^{y}$ denote vertex with index $i$ and its $y$ coordinate value, respectively. And $y_m$ denotes the y-axis threshold value in the decimeter unit. In our experiments, we set $y_m=-0.15$ and get a promising mouth region, resulting in $14379$ mouth indices. 
Afterward, we can utilize these lower mouth indices $\mathcal{I}$ to generate binary lower mouth mask vector $M \in \mathbb{R}^N$ for every 3D face model as follows:
\begin{equation}
    M_i= \begin{cases}1, i \in \mathcal{I}  \\ 0, \text{otherwise}\end{cases}.
\end{equation}

\noindent \textbf{Training Details}
Our framework is implemented by PyTorch and a differentiable computer vision package named Kornia\footnote{https://github.com/kornia/kornia}. 
So in fact our framework can be trained end-to-end.
Adam optimizer is adapted for all experiments with the initial learning rate of $1\times10^{-4}$. 
We train the A2EP module for 70 epochs with the same sequence length in each batch, 4s (100 frames) for the HDTF dataset, and 2s (50 frames) for the VoxCeleb2 dataset. 
For the TAFT module, the training process takes 15 epochs for convergence. 

The mouth weight $\lambda_{m}$ is set to 1.8, and the weights for photo-metric loss, perceptual loss, and style loss are set to $\lambda_{photo}=1.0$, $\lambda_{perc}=4.0$, and $\lambda_{s}=1000.0$ respectively.
All models are trained and tested on a single NVIDIA V100.

\noindent \textbf{Testing Details}
When testing, given arbitrary audio speech, the corresponding expression parameters sequence are firstly predicted by the A2EP. Note that here although we utilize one-hot vectors in training stage, to distinguish the different speaking styles instead of speakers' ids for facilitate the training process, following previous works, \textit{i.e.}, VOCA~\cite{cudeiro2019capture} and FaceFormer~\cite{faceformer2022}.
During inference, we can choose any one-hot vector to predict the expression parameters in A2EP, and then substitute the counterparts of 3DMM parameters extracted from the unseen target videos, obtaining the combined 3DMM parameters. 
%

After that, The TAFT module, empowered by the MAFB module, takes these combined parameters sequence and target video as input to synthesize the final talking face video. 
Note that when testing, we use the fixed reference image in the MAFB module to avoid generating video with illumination variations.

\begin{table}[t]
  \centering
  \footnotesize
  \caption{User study results of our model with state-of-the-arts. Positive scores indicate favorable comments from participants. Larger is better.}
  \label{tab:user_study}
  \begin{tabular}{l|ccc}
    \toprule
    \textbf{Methods} & \textbf{Lip-Sync} & \textbf{Image} & \textbf{Video} \\
    \midrule
    ATVG~\cite{chen2019hierarchical} & 1.12 & -0.83  & -1.13 \\
    Wav2Lip~\cite{prajwal2020lip} & 1.47 & 0.10 & 1.12 \\
    MakeItTalk~\cite{MakeItTalk} & -0.18 & -0.87 & -0.97 \\
    PC-AVS~\cite{zhou2021pose} & 0.83 & 1.47 & 0.99 \\
    \hline
    Ground Truth & 1.78 & 1.83 & 1.78 \\
    Ours & \textbf{1.83} & \textbf{1.78} & \textbf{1.61} \\
  \bottomrule
\end{tabular}
\end{table}

\begin{table}[t]
\centering
\caption{Quantitative comparison with video-based methods.}
\resizebox{\linewidth}{!}{
\begin{tabular}{cccccc}
\hline  
\textbf{Methods}& \textbf{LMD} $\downarrow$ & \textbf{SSIM} $\uparrow$ & \textbf{PSNR}$\uparrow$ & \textbf{CPBD}$\uparrow$ & \textbf{AV conf.}$\uparrow$\\
\hline  
NVP & 3.215 & 0.688 & 35.164 & 0.182 & 6.486 \\
FACIAL & 2.352 & 0.954 & 38.102 & 0.214 & 6.862 \\
 \hline
Ours & \textbf{1.423} & \textbf{0.965} & \textbf{40.702}& \textbf{0.246} & \textbf{7.985}\\
\hline 
\end{tabular}}
\label{tab:video_based_comparison}
\vspace{-0.4cm}
\end{table}

\subsection{Comparison with State-of-the-Arts}

\noindent \textbf{Quantitative Comparison.} We compare our method with recent state-of-the-art methods, including ATVG~\cite{chen2019hierarchical}, Wav2Lip~\cite{prajwal2020lip}, MakeItTalk~\cite{MakeItTalk}, and PC-AVS~\cite{zhou2021pose}. All the results are generated by using their released code. 
As shown in Tab.~\ref{tab:quan}, our method could achieve the best performance under most of the evaluation metrics on VoxCeleb and HDTF datasets, except the AV confidence metric. 
However, it is worthy pointing out that Wav2Lip is biased to do particularly well on the AC confidence metric, even better than ground truth for the HDTF dataset in Tab.~\ref{tab:quan}. It is because that Wav2Lip leverages an extra SyncNet architecture as a lip-sync loss during training, as revealed in~\cite{Lahiri_2021_LipSync3D}.
For the evaluation of synchronization between lip motion and the input audio, our method can achieve a very low score in the LMD metric, meaning that our method can generate a realistic lip movement as close as the original speaker.
The higher scores in the metrics SSIM, PSNR, and CPBD also indicate that the quality of our generated video is superior, which is important for the generated video to be realistic and vivid. 

Since almost all existing video-based methods, \textit{e.g.}, NVP~\cite{thies2020NVP} and FACIAL~\cite{zhang2021facial} are designed for specific speakers.
So it is infeasible and time-consuming to conduct a fair comparison in the large HDTF and VoxCeleb datasets since we need to re-train individual models for each speaker.
In order to compare with them, we conduct additional experiments on 3 non-intentional selection speakers against these video-based methods. The results are shown in Tab.~\ref{tab:video_based_comparison}.
From the results, we can find out that our method outperforms these video-based methods significantly even though they cost more training time and require individual models for each speaker.

\noindent \textbf{Qualitative Results.}
We also qualitatively compare our method with these methods as shown in Fig.~\ref{fig:quali}.
Among them, ATVG and PC-AVS can only generate cropped face images and the image resolution are much lower than ours
%
Although Wav2Lip~\cite{prajwal2020lip} and MakeItTalk~\cite{MakeItTalk} could generate the full images containing backgrounds, their synthetic faces with noticeably blurry mouth areas and inaccurate lip movements.
%
%
In comparison, our method could generate accurate and smooth lip motions with the 3DMM intermediate representation (see the second line from the bottom in Fig.~\ref{fig:quali}). 
Then, with the help of the target adaptive face translation module, we can generate photo-realistic talking face images which can inherit these correct lips performance while retaining the details in other regions. 
What's more, our generated results have the natural head pose without introducing any distortions and artifacts. 
On the other hand, our method could be generalized to different speakers without being re-trained or fine-tuned.
More results can be viewed in our supplementary materials.

\begin{figure}[t]
 \begin{center}
  \includegraphics[width=\linewidth]{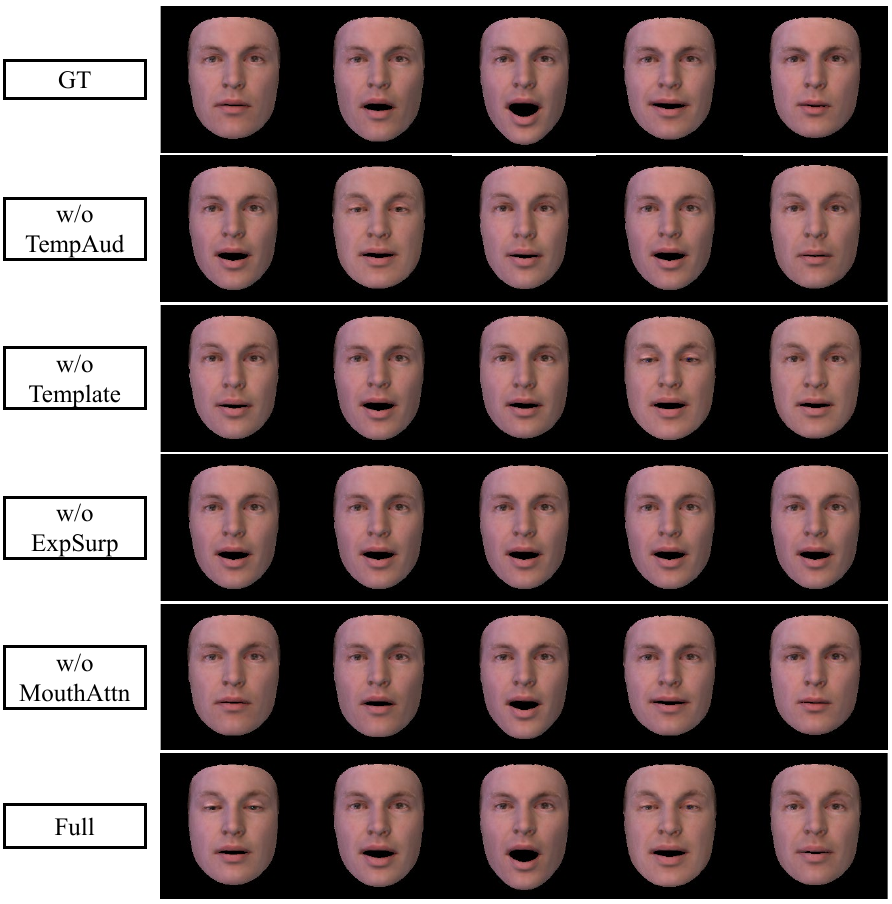}
 \end{center}
 \caption{Visualization results for the ablation study on A2EP module. Note that here we remove the head pose and identity information when rendering these images to focus on the dynamic lip movements.}
 \label{fig:ablAudio}
\end{figure}

\noindent \textbf{User Study.} We also conducted a user study to investigate the subjective evaluation of these synthesized talking face videos from different methods. 
We prepare 10 videos from VoxCeleb and 10 videos from the HDTF testing dataset to generate 20 talking face videos in total.
For a fair comparison, we shuffle the generated videos and invite 18 volunteers to rate these videos from three aspects: 1) audio-lip synchronization, 2) photo-realistic image quality, and 3) entire video realism. 
The evaluation scores are divided into five degrees between -2  (very bad) to 2 (very good). 
After collecting all evaluation results, we calculate the average scores for each method as listed in Tab.~\ref{tab:user_study}. 
From the results, we can easily find that almost all participants favor our generated results over other methods.

\begin{table}[t]
\centering
\caption{Ablations. Quantitative analysis of ablating key components in our method on part of the HDTF dataset. ``-" denotes it's hard to compute this metric.}
\label{tab:ablations}
\resizebox{\linewidth}{!}{
\begin{tabular}{cccccc}
\hline  
\textbf{Models}& \textbf{LMD} $\downarrow$ & \textbf{SSIM} $\uparrow$ & \textbf{PSNR}$\uparrow$ & \textbf{CPBD}\textbf{}$\uparrow$ & \textbf{AV conf.}$\uparrow$\\
\hline  
w/o TempAud & 2.826 & 0.802 & 31.120 & 0.153 & 5.830 \\
w/o ExpSurp & 2.432 & 0.813 & 30.824 & 0.138 & 5.542 \\
w/o Template & 1.582 & 0.820 & 31.421 & 0.148 & 6.183 \\
w/o MouthAttn & 1.158 & 0.894 & 31.240 & 0.156 & 6.581 \\
w/o MAFB & - & 0.430 & 32.082 & 0.136 & - \\
w/o MorphoAug & 1.263 & 0.782 & 35.128 & 0.183 & 6.216 \\
\hline
Full model & \textbf{0.952} & \textbf{0.914} & \textbf{37.212} & \textbf{0.198} & \textbf{6.932} \\
\hline 
\end{tabular}}
\label{video_based_comparison}
\end{table}
\vspace{-0.8em}

\subsection{Ablation Study}

To evaluate the effectiveness of each proposed component, we conduct an ablation study on part of the HDTF dataset. And the quantitative results are shown in Tab.~\ref{tab:ablations}.

\noindent \textbf{Ablation on A2EP.} We design four variants to validate critical components in this module: (1) use pre-trained DeepSpeech network~\cite{zhang2021facial, wu2021imitating} to replace our temporal-aware audio encoder (\textbf{w/o TempAud)}, (2) remove the explicit supervision in the fine-grained 3D facial vertices aspects (\textbf{w/o ExpSurp}), (3) supervise the expression in the facial vertices aspect, but do not use the template face to calculate adjusted ground truth and our prediction (\textbf{w/o Template}), (4) remove the larger weight on lower mouth region when computing the prediction loss (\textbf{w/o MouthAttn}).
For each variant, we obtain the predicted expression with several driven audios and render the 3DMM parameters for comparison (see Fig.~\ref{fig:ablAudio}). From our experiments, the models \textbf{w/o TempAud} and \textbf{w/o ExpSurp} always obtain poor prediction performance that the talking mouth keeps open all the time. Whereas, the model \textbf{w/o Template} could generate results with subtle motions and have temporal jitter effects. The lip shapes are correct in \textbf{w/o MouthAttn} model but are not vivid compared to our full model.

\begin{figure}[t]
 \begin{center}
  \includegraphics[width=0.85\linewidth]{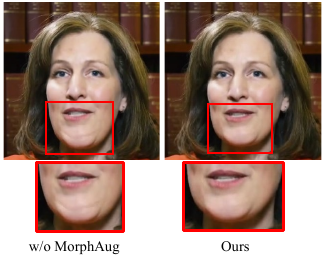}
 \end{center}
 \vspace{-.4cm}
 \caption{Visualization results for the ablation study on Target Adaptive Face Translation module.}
 \label{fig:ablFaceBlend}
\end{figure}

\noindent \textbf{Ablation on TAFT.}
We conduct two ablation variants for the TAFT module: (1) remove the MAFB module, instead, we directly use the rendered face and reference image as the generator input (\textbf{w/o MAFB}), and (2) remove the morphology-based mask augmentation in  MAFB module (\textbf{w/o MorphoAug)}.
The image quality drops dramatically when removing the face blending module. In fact, in this case, the generator always fails to translate the rendered faces into corresponding photo-realistic face images. 
As illustrated in Fig.~\ref{fig:ablFaceBlend}, when removing the morphology-based mask augmentation operations in the MAFB module, it tends to obtain unnatural chins since the rendered face shape may be smaller or larger than the target one. 

\noindent \textbf{Key to Generalizability.}
The generalizability to unseen identities is achieved by our TAFT module and the disentanglement ability of 3DMM.
Since it is hard to model the complex image backgrounds, previous works need re-training to fit the model to new backgrounds of unseen videos~\cite{yi2020audio_personalized, zhang2021facial}
By contrast, we blend the fine-grained lip-synchronized rendered image into the target video directly using the TAFT module, greatly easing the learning burden of the network.
While the A2EP aims at generating audio-related expression parameters, which are independent to the id-related appearance.

\section{Ethical Considerations}

Our method focuses on synthesizing vivid talking face videos, which is intended for developing digital entertainment and an advanced teleconferencing system. 
However, it may also be misused maliciously on social media, resulting in negative impacts (\textit{e.g.}, 
spread of fake news).
Nevertheless, we believe studying these methods is important and necessary to help develop robust and powerful Deepfake detection algorithms~\cite{rossler2019faceforensics}. 
And we also should admit that they have promising applications to enhance our daily life, for example, visual dubbing and virtual avatars.

\section{Conclusion and Discussions}

We introduce GSmoothFace, a simple yet effective talking face generation framework via making full use of the fine-grained 3D face model.
On one hand, we consider the fact that the limitations of existing 3D face reconstruction algorithms and propose explicitly supervising the expression parameters in dense facial vertices aspect.
And in this case, we have the flexibility to attach more weights on the speech-related areas, \textit{e.g.} the mouth region. 
On the other hand, the carefully designed TAFT module leverages the rendered 3D face and binary facial mask image from the 3DMM model without further effort, resulting in synthesizing the photo-realistic talking face videos with few artifacts.
%
%
%
Our method generalized to unseen identities without re-training, owing to the well-designed TAFT module.
Extensive experimental results validate the effectiveness of our method.
	
{{\section*{Acknowledgment}} This work was supported in part by Shenzhen General Program No.JCYJ20220530143600001, by the Basic Research Project No.HZQB-KCZYZ-2021067 of Hetao Shenzhen HK S\&T Cooperation Zone, by Shenzhen Hong Kong Joint Funding No.SGDX20211123112401002, by Shenzhen Outstanding Talents Training Fund, by Guangdong Research Project No.2017ZT07X152 and No.2019CX01X104, by the Guangdong Provincial Key Laboratory of Future Networks of Intelligence (Grant No.2022B1212010001), The Chinese University of Hong Kong, Shenzhen, by the NSFC 61931024\&81922046, by Tencent Open Fund.}

	
	%

	


	\ifCLASSOPTIONcaptionsoff
	\newpage
	\fi

	
	{
		\bibliographystyle{IEEEtran}
		\bibliography{IEEEabrv,IEEEexample}
	}

\begin{IEEEbiography}[{\includegraphics[width=1in,height=1.25in,clip,keepaspectratio]{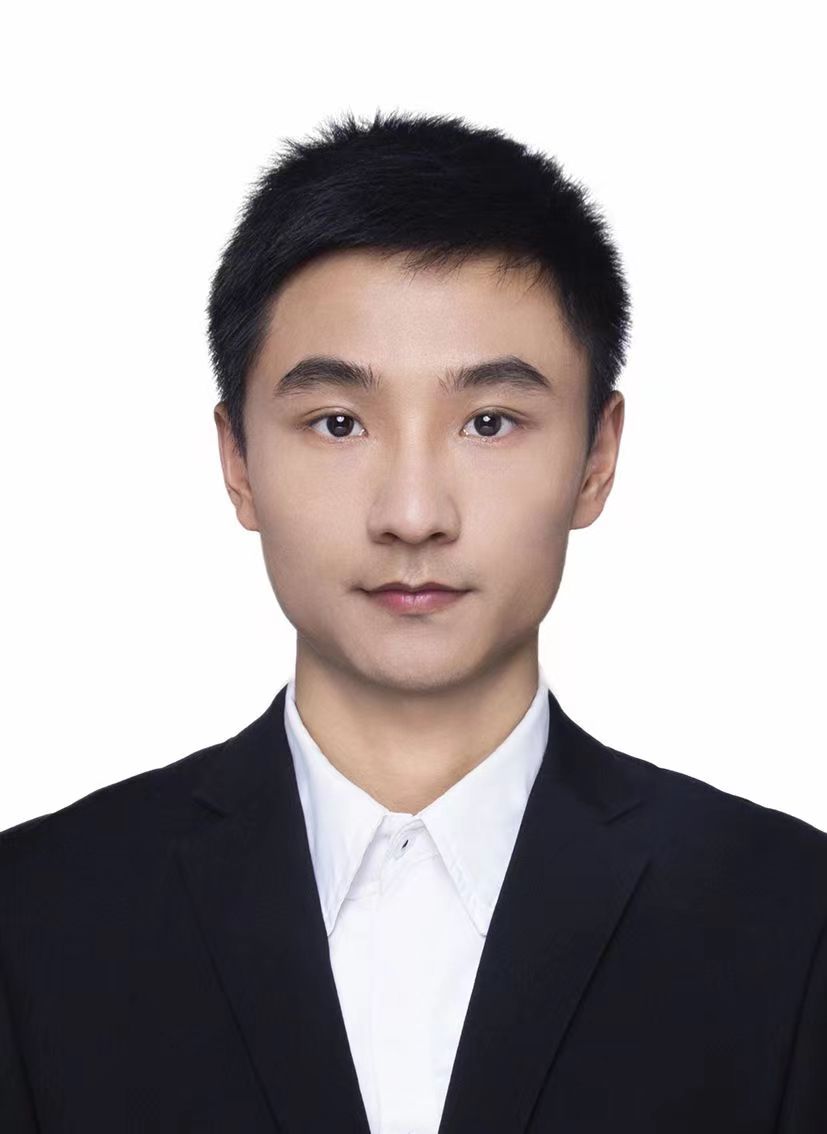}}]{Haiming Zhang} received the B.S. and M.S. degrees from the Beijing Institute of Technology. He is currently pursuing a Ph.D. degree at the Chinese University of Hong Kong (Shenzhen). His research interests focus on 3D computer vision, especially 3D point cloud analysis, and generative models. He has published top conferences or journal papers, such as CVPR, ACMMM, TPAMI, etc.
\end{IEEEbiography}

\begin{IEEEbiography}[{\includegraphics[width=1in,height=1.25in,clip,keepaspectratio]{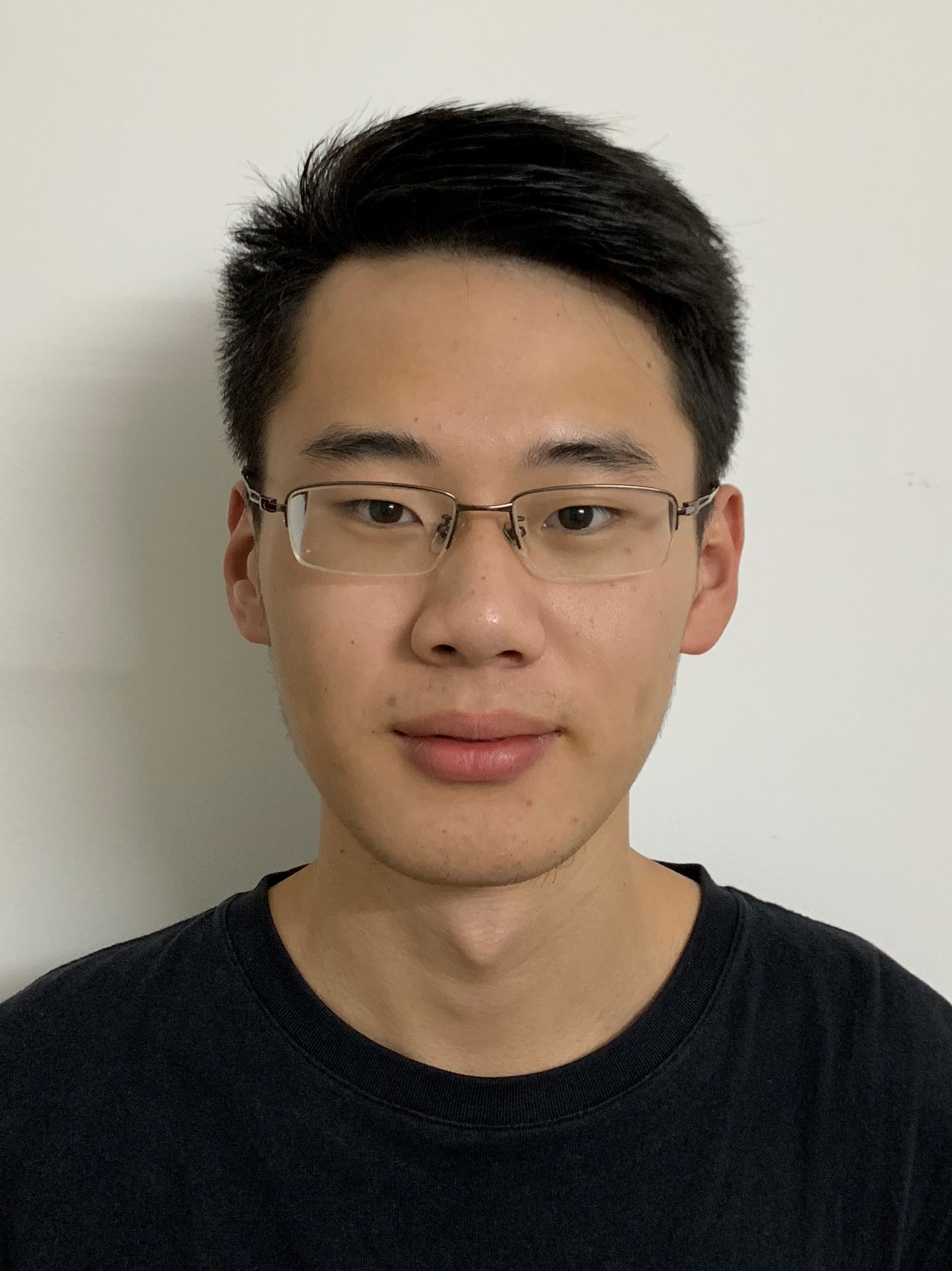}}]{Zhihao Yuan}
received the B.E. degree in Software Engineering from the School of Data and Computer Science, Sun Yat-sen University, Guangzhou, China in 2020. He is currently pursuing the Ph.D. degree with the Deep Bit Lab, Future Network of Intelligence Institute (FNii), Chinese University of Hong Kong, Shenzhen, China. His research interests include 3D computer vision and language.
\end{IEEEbiography}

\begin{IEEEbiography}[{\includegraphics[width=1in,height=1.25in,clip,keepaspectratio]{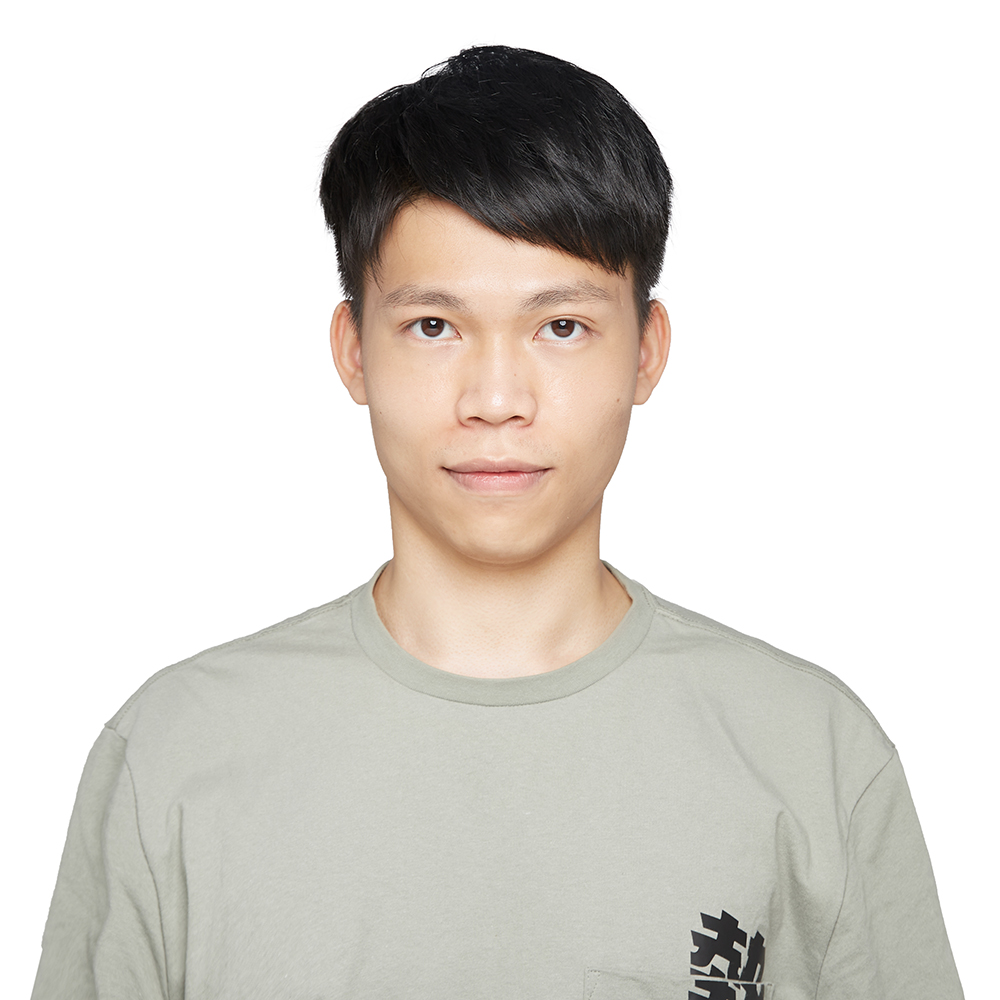}}]{Chaoda Zheng} received the B.Eng. and M.Eng. degrees from the South China University of Technology. He is currently pursuing a Ph.D. degree at the Chinese University of Hong Kong (Shenzhen). His research interests focus on 3D computer vision, especially 3D point cloud analysis, in the field of which he has published multiple top conferences or journal papers, such as CVPR, ICCV, ECCV, NeurIPS, TIP, etc.
\end{IEEEbiography}

\begin{IEEEbiography}[{\includegraphics[width=1in,height=1.25in,clip,keepaspectratio]{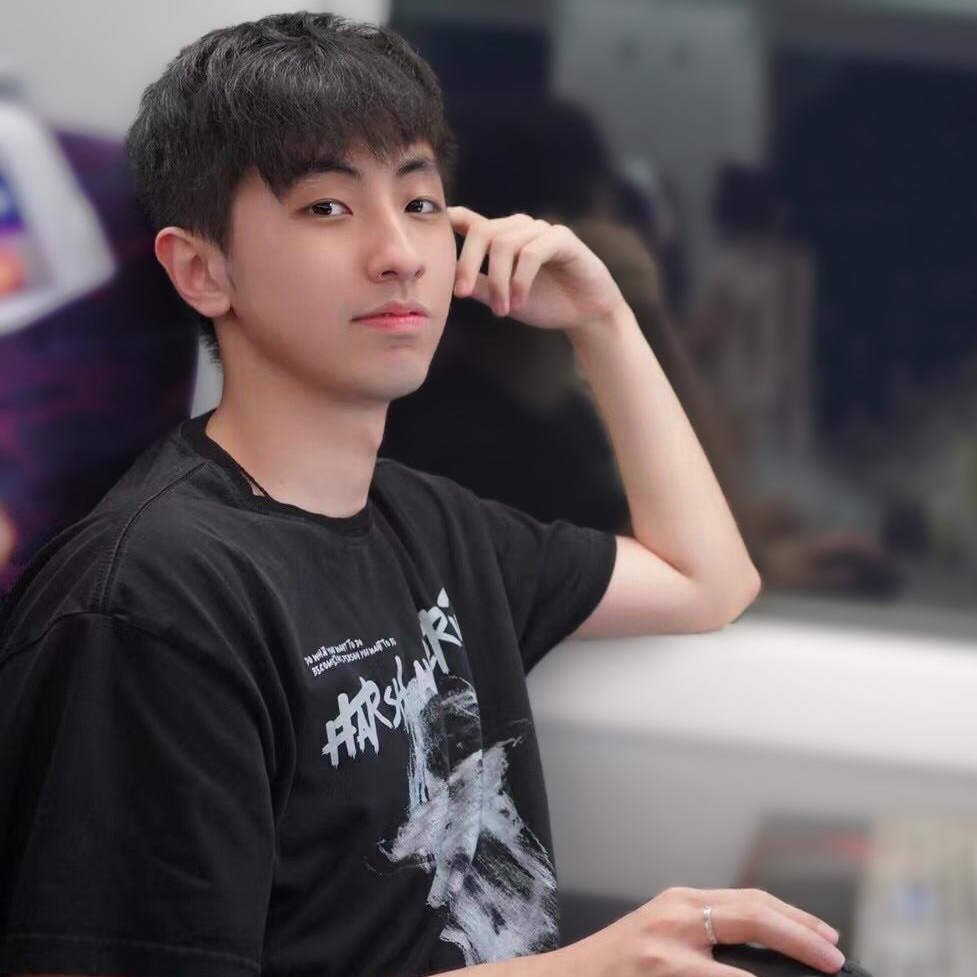}}]{Xu Yan} received the B.S. degree from the San Yat-sen University and Master degree from the Chinese University of Hong Kong (Shenzhen). He is currently pursuing a Ph.D. degree at the Chinese University of Hong Kong (Shenzhen). His research interests focus on 3D computer vision, especially 3D point cloud analysis. He has published more than 10 papers in top conferences, such as CVPR, ICCV, ECCV, NeurIPS, AAAI, IJCAI, ISBI, etc.
\end{IEEEbiography}

 \begin{IEEEbiography}[{\includegraphics[width=1in,height=1.25in,clip,keepaspectratio]{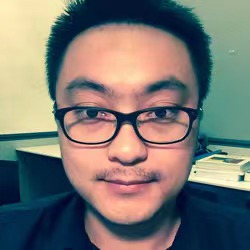}}]{Baoyuan Wang} is the VP of engineering at Xiaobing.ai, a startup spun off from Microsoft in 2020. Prior to this, he was a Sr.Principal Researcher and Manager at Microsoft AI platform and HoloLen Mix-reality team from 2015 to 2021, and a lead researcher at MSRA from 2012 to 2015. His research interests include computational photography, digital human synthesis, and conversational AI. He got both his Bechler and Ph.D. degrees in computer science at Zhejiang University in 2007 and 2012 respectively. He was an engineering intern at Infosys, Mysore, India from 2006 to 2007 and a research intern at Microsoft from 2009 to 2012.
\end{IEEEbiography}

\begin{IEEEbiography}[{\includegraphics[width=1in,height=1.25in,clip,keepaspectratio]{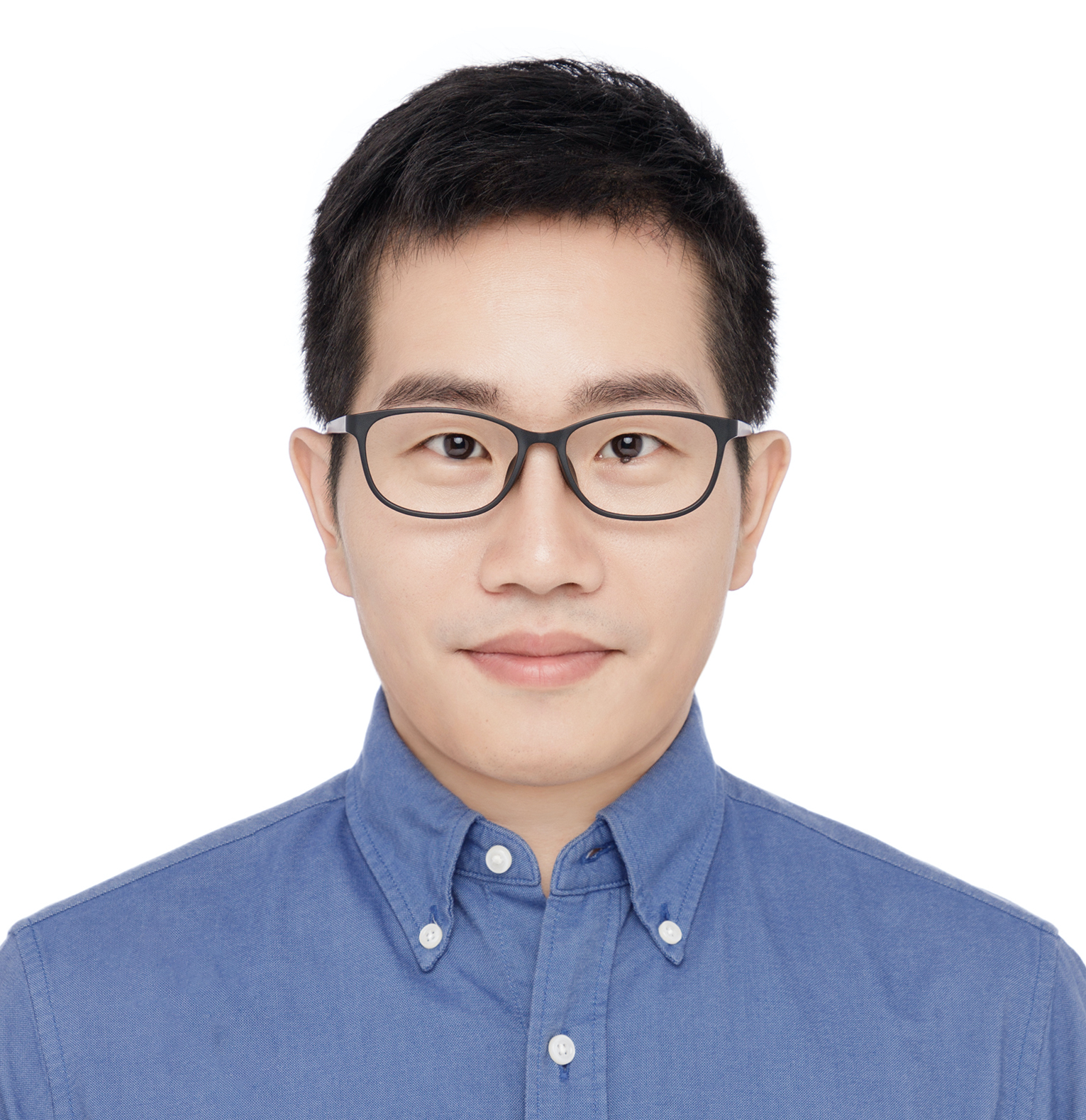}}]{Guanbin Li} is currently an associate professor in School of Data and Computer Science, Sun Yat-sen University. He received his PhD degree from the University of Hong Kong in 2016. His current research interests include computer vision, image processing, and deep learning. He is a recipient of ICCV 2019 Best Paper Nomination Award. He has authorized and co-authorized on more than 100 papers in top-tier academic journals and conferences. He serves as an area chair for the conference of VISAPP. He has been serving as a reviewer for numerous academic journals and conferences such as TPAMI, IJCV, TIP, TMM, TCyb, CVPR, ICCV, ECCV and NeurIPS.
\end{IEEEbiography}

\begin{IEEEbiography}[{\includegraphics[width=1in,height=1.25in,clip,keepaspectratio]{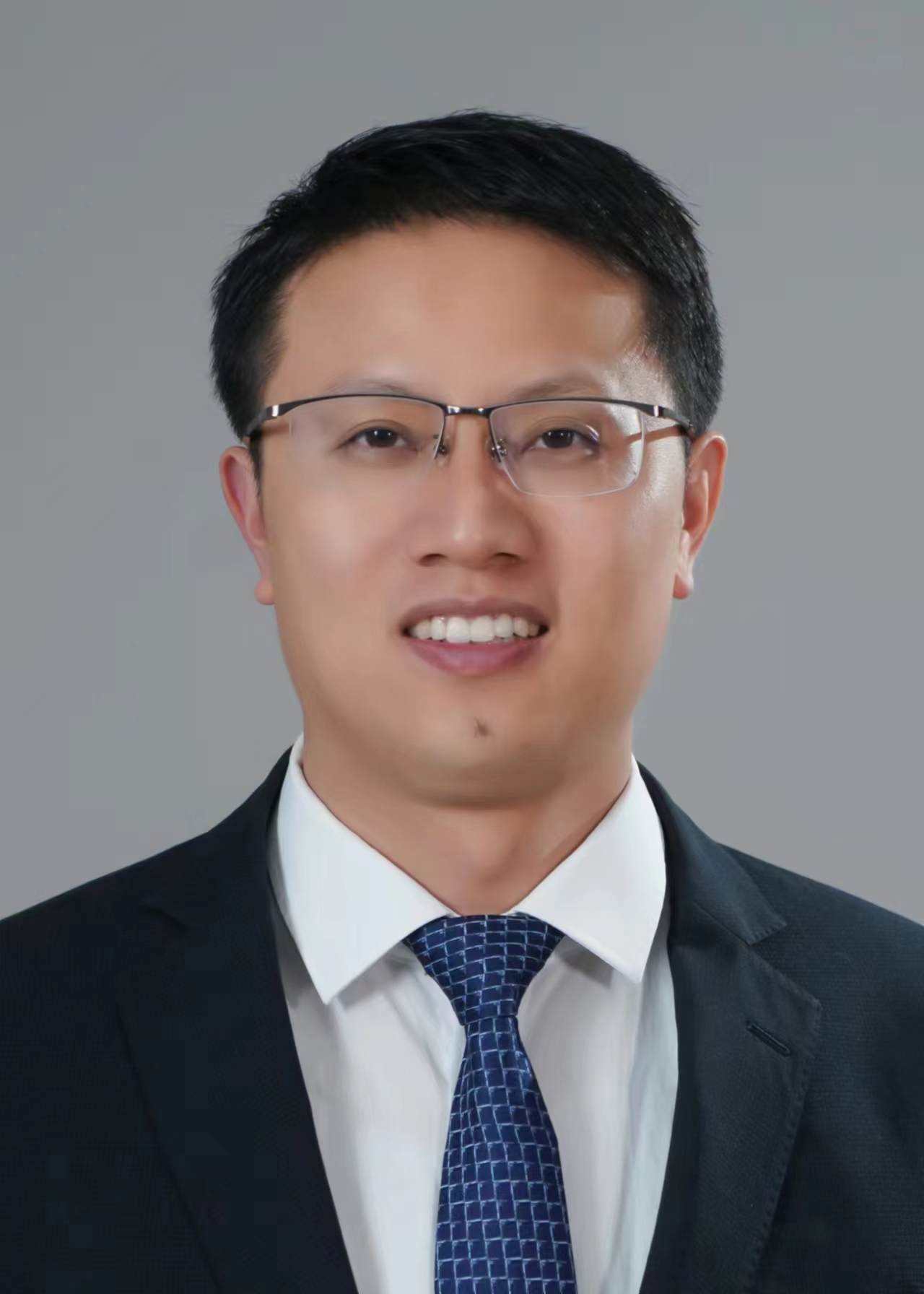}}]{Song Wu} works as the professor and doctoral supervisor at Shenzhen University.  His current research interests focus on the major clinical needs of non-invasive early detection of bladder cancer, and the scientific problem of how to obtain early cancer signals in a complex microenvironment. A series of work on new methods of early signal acquisition of bladder cancer and new paradigm of non-invasive detection of urine. He has hosted multiple national level scientific research projects, including Led key projects of the National Natural Science Foundation of China, the Excellent Youth Fund of the National Natural Science Foundation of China, and the Key Research and Development Special Project of the Ministry of Science and Technology. Carried out research and transformation work. Transferred 4 patents, developed and launched 2 products, obtained 2 national medical device certificates, and provided nearly 10 million screening services for non-invasive urine early screening products. Editor in chief of two monographs, "Introduction to Precision Medicine" and "Say Goodbye to Urinary Tract Stones", and served as the editorial board member of Wu Jieping's Urology and bladder cancer Diagnosis and Treatment Guide, the classic treatise of Chinese urology. Supported by the National Excellent Youth Science Fund, and awarded awards such as the Guangdong Provincial Youth Science and Technology Award, Shenzhen Youth Science and Technology Award, Guangdong Provincial Science and Technology Innovation Leading Talent, and National May Day Labor Medal
\end{IEEEbiography}

\begin{IEEEbiography}[{\includegraphics[width=1in,height=1.25in,clip,keepaspectratio]{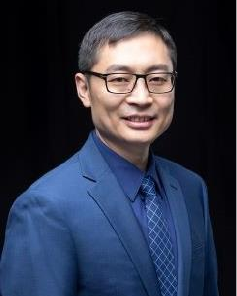}}]{Shuguang Cui} (Fellow, IEEE) received the Ph.D. degree in electrical engineering from Stanford University, California, USA, in 2005. Afterwards, he has been working as Assistant, Associate, Full, Chair Professor in electrical and computer engineering
with the University of Arizona, Texas A\&M University, UC Davis, and the Chinese University of Hong Kon, Shenzhen (CUHKSZ), respectively. He has also been the Executive Dean of the School of Science and Engineering, CUHKSZ, and the Executive Vice Director at the Shenzhen Research Institute of Big Data. His current research interests focus on data driven large-scale system control and resource management, large data set analysis, the IoT system design, energy harvesting based communication system design, and cognitive network optimization. He was selected as the Thomson Reuters Highly Cited Researcher and listed in the Worlds’ Most Influential Scientific Minds by ScienceWatch in 2014. He was a recipient of the IEEE Signal Processing Society 2012 Best Paper
Award. He has served as the general co-chair and TPC co-chairs for many IEEE conferences. He has also been serving as the Area Editor of the IEEE Signal Processing Magazine, and an Associate Editor of the IEEE Transactions on Big Data, IEEE Transactions on Signal Processing, IEEE JSAC Series on Green Communications and Networking, and IEEE Transactions on Wireless Communications. He has bee the elected member for IEEE Signal Processing Society SPCOM Technical Committee (2009–2014) and the elected Chair for IEEE ComSoc Wireless Technical Committee (2017–2018). He is a member of the Steering Committee of the IEEE TRANSACTIONS ON BIG DATA and the Chair of the Steering Committee of the IEEE TRANSACTIONS ON COGNITIVE COMMUNICATIONS
AND NETWORKING. He was also a member of the IEEE ComSoc Emerging Technology Committee. He was elected as an IEEE ComSoc Distinguished Lecturer in 2014, and IEEE VT Society Distinguished Lecturer in 2019. He has won the IEEE ICC best paper award, ICIP best paper finalist, the First Class Prize in Natural Science from Chinese Institute of Electronics, and the First Class Prize in Technology Invention from the China Institute of Communications all in 2020.
\end{IEEEbiography}

\begin{IEEEbiography}[{\includegraphics[width=1in,height=1.25in,clip,keepaspectratio]{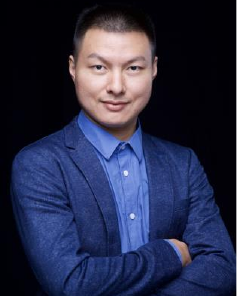}}]{Zhen Li}
 is currently serving as an assistant professor in the School of Science and Engineering as well as a research scientist in Shenzhen Research Institute of Big data, the Chinese University of Hong Kong, Shenzhen (CUHKSZ). He received his Ph. D. degree in Computer Science from University of Hong Kong (2014-2018), his master degree in Communication and Information Systems from Sun Yat-sen University (2011-2014) and his bachelor degree in Automation from Sun Yat-sen University (2007-2011). He also worked as a visiting scholar in University of Chicago in 2018 and a visiting student in Toyota Technological Institute at Chicago (TTIC) in 2016. His research interests include medical imaging and medical big data, AI interdisciplinary research and computer vision. He has published many papers in top conferences and journals, such as MICCAI, ISBI, CVPR, ICCV, ECCV, IJCAI, AAAI, ECAI, RECOMB, Nature Communications, PLOS CB and Cell Systems, etc. Meanwhile, he was the core team member for the champion of the 12th Critical Assessment of Protein Structure Prediction (CASP12), with the published paper receiving the PLOS CB 2018 Breakthrough and Innovation Awards and being Web-of-Science Highly Cited paper.
\end{IEEEbiography}
 
\end{document}